\documentclass[conference]{IEEEtran}
\IEEEoverridecommandlockouts
\usepackage{cite}
\usepackage{amsmath,amssymb,amsfonts}
\usepackage{algorithmic}
\usepackage{graphicx}
\usepackage{textcomp}
\usepackage{xcolor}
\usepackage{multirow}
\usepackage[super]{nth}
\def\BibTeX{{\rm B\kern-.05em{\sc i\kern-.025em b}\kern-.08em
    T\kern-.1667em\lower.7ex\hbox{E}\kern-.125emX}}

\makeatletter
\newcommand{\linebreakand}{%
  \end{@IEEEauthorhalign}
  \hfill\mbox{}\par
  \mbox{}\hfill\begin{@IEEEauthorhalign}
}
\makeatother

\begin{document}

\title{Multi-Signal Reconstruction Using Masked Autoencoder From EEG During Polysomnography 
\footnote{{\thanks{This work was supported by the Institute of Information \& Communications Technology Planning \& Evaluation (IITP) grant, funded by the Korea government (MSIT) (No. 2019-0-00079, Artificial Intelligence Graduate School Program (Korea University)) and was partly supported by the Institute of Information \& communications Technology Planning \& Evaluation (IITP) grant, funded by the Korea government (MSIT) (No. 2021-0-02068, Artificial Intelligence Innovation Hub).}
}}
}

\author{

\IEEEauthorblockN{Young-Seok Kweon}
\IEEEauthorblockA{\textit{Dept. of Brain and Cognitive Engineering} \\
\textit{Korea University}\\
Seoul, Republic of Korea \\
youngseokkweon@korea.ac.kr}

\and

\IEEEauthorblockN{Gi-Hwan Shin}
\IEEEauthorblockA{\textit{Dept. of Brain and Cognitive Engineering} \\
\textit{Korea University} \\
Seoul, Republic of Korea \\
gh\_shin@korea.ac.kr}

\linebreakand 

\IEEEauthorblockN{Heon-Gyu Kwak}
\IEEEauthorblockA{\textit{Dept. of Artificial Intelligence} \\
\textit{Korea University} \\
Seoul, Republic of Korea \\
hg\_kwak@korea.ac.kr} 

\and

\IEEEauthorblockN{Ha-Na Jo}
\IEEEauthorblockA{\textit{Dept. of Artificial Intelligence} \\
\textit{Korea University} \\
Seoul, Republic of Korea \\
hn\_jo@korea.ac.kr} 

\and

\IEEEauthorblockN{Seong-Whan Lee}
\IEEEauthorblockA{\textit{Dept. of Artificial Intelligence} \\
\textit{Korea University} \\
Seoul, Republic of Korea \\
sw\_lee@korea.ac.kr}

}

\maketitle

\begin{abstract}
Polysomnography (PSG) is an indispensable diagnostic tool in sleep medicine, essential for identifying various sleep disorders. By capturing physiological signals, including EEG, EOG, EMG, and cardiorespiratory metrics, PSG presents a patient's sleep architecture. However, its dependency on complex equipment and expertise confines its use to specialized clinical settings. Addressing these limitations, our study aims to perform PSG by developing a system that requires only a single EEG measurement. We propose a novel system capable of reconstructing multi-signal PSG from a single-channel EEG based on a masked autoencoder. The masked autoencoder was trained and evaluated using the Sleep-EDF-20 dataset, with mean squared error as the metric for assessing the similarity between original and reconstructed signals. The model demonstrated proficiency in reconstructing multi-signal data. Our results present promise for the development of more accessible and long-term sleep monitoring systems. This suggests the expansion of PSG's applicability, enabling its use beyond the confines of clinics. 
\end{abstract}

\begin{small}
\end{small}

\section{INTRODUCTION}
Polysomnography (PSG) is a comprehensive recording of the physical changes that occur during sleep \cite{psg}. As the cornerstone of sleep medicine, PSG is utilized extensively to diagnose sleep disorders, including sleep apnea, periodic limb movement disorder, narcolepsy, and rapid eye movement (REM) behavior disorder \cite{narcolepsy, rbd, periodicleg}. This multifaceted diagnostic tool simultaneously captures an array of physiological signals such as electroencephalography (EEG), electrooculography (EOG), electromyography (EMG), electrocardiography (ECG), and respiratory metrics like heart rate and breathing patterns. The integration of these signals provides the patient's sleep architecture, including the transitions between different sleep stages, the quality of sleep, and the presence of sleep-related anomalies. Given the complexity and richness of the data acquired, PSG serves as a valuable source for both clinical assessments and research into the mechanisms and qualities of sleep\cite{ssqpsg, narcolepsy, rbd}. Despite its comprehensive nature, PSG's reliance on specialized equipment and expert analysis limits its use to predominantly clinical settings, often for severe sleep disorders. The necessity for expertise and equipment restricts the feasibility of long-term or home-based monitoring. Addressing these limitations, our study aims to perform PSG by developing a system that requires only a single EEG measurement.

Deep learning is an effective tool for analyzing bio-signals \cite{ref1, kweon2021automatic, ref9}, classifying the intention of humans from EEG \cite{ref2, kweon2023development, ref10}, and identifying medical states \cite{kweon2020prediction, ref3}. A masked autoencoder is a type of deep learning model that is designed for the task of unsupervised learning \cite{maskedae}. It operates by intentionally obscuring a part of its input data and then learning to predict the missing parts. This self-supervised approach enables the model to capture the underlying structure of the data in a way that is particularly useful for reconstruction tasks, and understanding data structure is important to extract features of EEG \cite{ref4, ref6}. The strength of a masked autoencoder lies in its ability to handle and predict complex patterns within large and noisy datasets, making it an invaluable tool for signal processing and data restoration \cite{zhang2022mask}. 

\begin{figure*} 
\begin{center}
\includegraphics[width=1\linewidth]{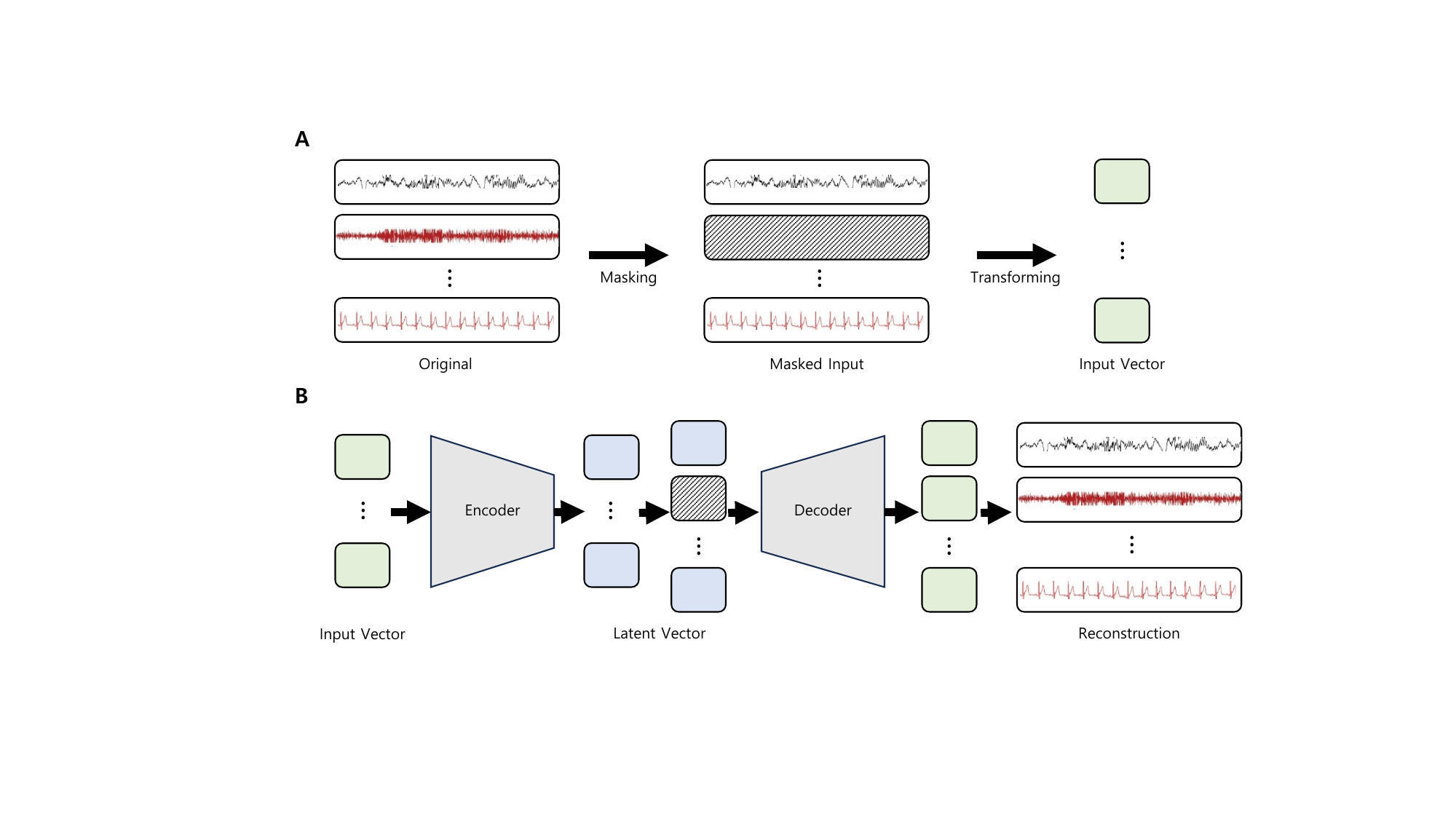}
\end{center}
\caption{Overall framework of masked autoencoder from masking and transforming signals (A) to multi-signal reconstruction (B).}
\label{fig:framework}
\end{figure*}

In this study, we designed a masked autoencoder with the capability to reconstruct multi-signals using single-channel EEG data collected via PSG. Our model was trained and tested by the sleep-EDF-20 dataset. To ensure the precision of our reconstructed signals, we employed the mean squared error (MSE) as a metric to evaluate the similarity between the original and the autoencoder's reconstructed outputs. Our results present the potential to more simply diagnose sleep disorders and expand the applicability of PSG in a variety of settings by reconstructing a multi-signal from a single signal.

\section{MATERIALS AND METHODS}

\subsection{Dataset}
In this study, we utilized the publicly available Sleep-EDF-20 dataset to train and evaluate our masked autoencoder model \cite{sleepedf}. The Sleep-EDF dataset comprises PSG recordings and accompanying hypnograms from 78 healthy Caucasian subjects, with an age range of 25 to 101 years. To ensure consistency and comparability with extant studies, we selected a subset of 20 subjects for inclusion in our analysis. The dataset was handled by the same process as described in previous EEG studies \cite{ref7, jeong20222020}. This preprocessing included signal normalization, artifact removal, and segmentation into epochs \cite{ref5}.

\subsection{Masked Autoencoder}
Our study designed a masked autoencoder framework for the reconstruction of multi-signals from single-channel EEG data, as illustrated in the accompanying Fig. \ref{fig:framework}. The framework operates in two main stages: masking and transforming signals and multi-signal reconstruction.

The initial stage involved the masking of the original signals. As depicted, the original signals underwent a masking process where certain parts of the data were intentionally obscured. This manipulated data, referred to as the `Masked Input', was then transformed into an input vector that encapsulates essential features and information necessary for the reconstruction process.

The transformed input vector was subsequently fed into the encoder-decoder architecture of the autoencoder. The encoder compressed the input into a latent vector, capturing the intrinsic data representations. This latent vector was then processed by the decoder, which reconstructs the multi-signal data. The reconstructed output was a regenerated version of the original signals, produced by the masked autoencoder. We performed cosine similarity as a loss to train our masked autoencoder.

The entire framework is optimized to accurately reconstruct the original physiological signals from the masked inputs, thus facilitating the analysis of polysomnographic data using a single EEG channel.

\subsection{Evaluation}
We adopted the MSE as the primary metric for quantifying the similarity between the original and reconstructed signals. MSE is a widely used statistical measure that computes the average squared difference between the estimated values and what is estimated. It serves as a robust indicator of the precision of the reconstructed signals, offering a clear quantification of the reconstruction error. The lower the MSE, the closer the reconstructed signal is to the original, indicating a more accurate reconstruction by the masked autoencoder. We calculated the MSE for each reconstructed signal against its original counterpart and averaged them based on classes, such as wakefulness, non-REM stage 1 (N1), stage 2 (N2), stage 3 (N3), and REM. 

\section{RESULTS}

\begin{figure}[t] 
\begin{center}
\includegraphics[width=1.0\linewidth]{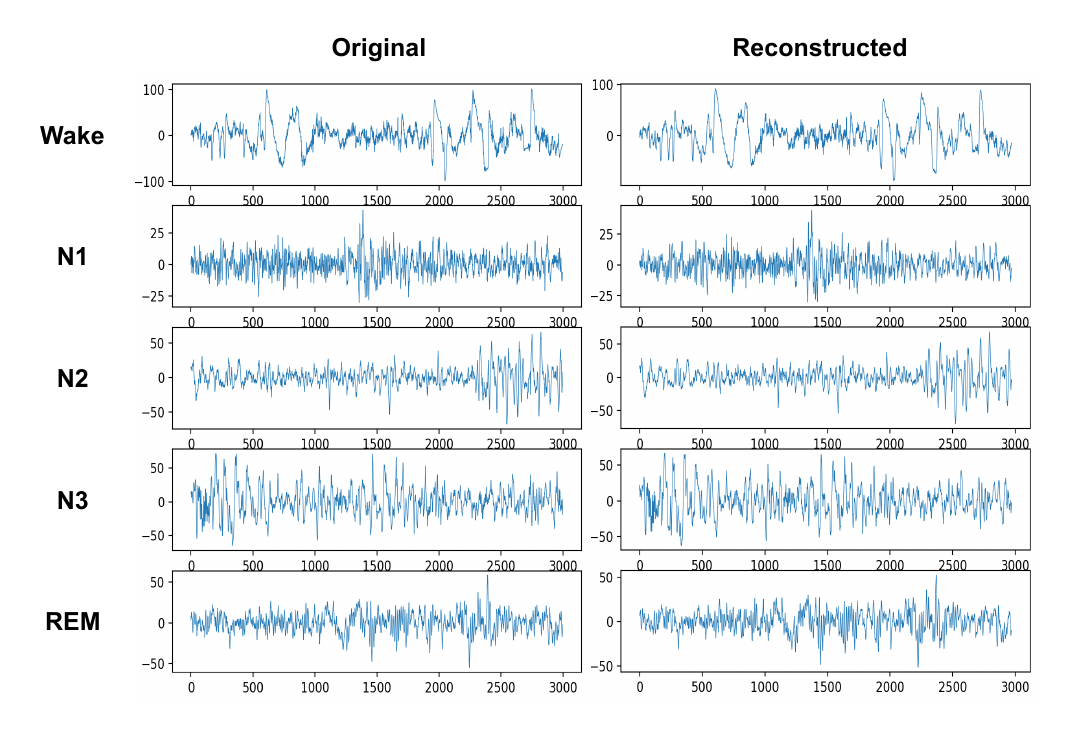}
\end{center}
\caption{Example of original and reconstructed EEG in Wake, non-rapid eye movement sleep stage 1 (N1), stage 2 (N2), stage 3 (N3), and rapid eye movement sleep (REM)}
\label{fig:fig2}
\end{figure}

Fig. \ref{fig:fig2} illustrated examples of original and reconstructed EEG signals across various sleep stages: Wakefulness, N1, N2, N3, and REM. In the wakefulness stage, the original EEG exhibited signals with relatively large amplitudes at certain instances. Our model effectively replicated these characteristics, reproducing similar amplitudes at corresponding times. In the N1 state, activation of theta waves and reduced amplitudes were prominent features, which were reflected in the reconstructed signals. For the N2 stage, a notable increase in amplitude at a specific moment, an essential indicator of potential sleep stage transitions, was accurately captured by our model. In the N3 stage, the original signals exhibited enhancing low-frequency components, a characteristic that was also well-replicated in the reconstructed signals. For the REM stage, our model discerningly reproduced the patterns similar to wakefulness but with attenuated amplitudes, ensuring clarity in distinguishing between the REM and wakefulness stages. This level of accuracy, which reproduces key features throughout the sleep stages, highlighted the effectiveness of our model in the reconstruction of EEG signals.

Table \ref{tab:my-table} presents the MSE between the original and reconstructed signals across various sleep stages: wake, N1, N2, N3, and REM for different inputs and reconstructions. In this study, two primary signals, EEG and EOG, were utilized as inputs to the model, setting a solid foundation for a comprehensive analysis and reconstruction process. For each kind of input signal, the model calculated the MSE against a diverse set of reconstructed signals, encompassing EEG (FPz-Cz), EEG (Pz-Oz), EMG, and EOG, with a deliberate exclusion of each signal's reconstruction against itself to prevent redundancy and enhance analytical clarity. When EEG is used as the input, the MSE between the original and reconstructed EOG signals during wake was 10.2, and during N1, it was 5.1. For the reconstructed EMG signal, the corresponding MSE values were 9.1, 5.0, 4.3, 4.4, and 6.9, respectively. For the reconstructed EEG (Pz-Oz) signal, the MSE values were 2.1, 2.3, 1.9, 2.0, and 2.2 across the sleep stages. When EOG was used as the input signal, the reconstructed EEG signals exhibited MSE values of 11.1, 5.9, 2.1, 3.4, and 5.0 for each sleep stage. For the reconstructed EMG signal, the MSE values were 8.4, 4.9, 4.8, 4.8, and 5.2, respectively. For the reconstructed EEG (Pz-Oz) signal, the MSE values across the sleep stages were 3.4, 3.3, 3.0, 3.1, and 2.9.

\begin{table}[t]
\centering
\caption{MSE between original and reconstructed signals.}
\label{tab:my-table}
\resizebox{\columnwidth}{!}{%
\begin{tabular}{c|c|ccccc}
\hline
Input & Reconstruction & Wake & N1  & N2  & N3  & REM \\ \hline
      & EOG            & 10.2 & 5.1 & 3.4 & 3.7 & 7.5 \\
EEG   & EMG            & 9.1  & 5.0 & 4.3 & 4.4 & 6.9 \\
(FPz-Cz) & EEG (Pz-Oz) & 2.1  & 2.3 & 1.9 & 2.0 & 2.2 \\ \hline
      & EEG (FPz-Cz)   & 11.1 & 5.9 & 2.1 & 3.4 & 5.0 \\
EOG   & EMG            & 8.4  & 6.1 & 4.9 & 4.8 & 5.2 \\
      & EEG (Pz-Oz)   & 3.4  & 3.3 & 3.0 & 3.1 & 2.9 \\ \hline
\end{tabular}%
}
\end{table}

\section{DISCUSSION}
We presented a masked autoencoder with the capability to reconstruct multi-signals using single-channel EEG data collected via PSG. Our model was trained and tested by the sleep-EDF-20 dataset. To ensure the precision of our reconstructed signals, we employed the MSE as a metric to evaluate the similarity between the original and the autoencoder's reconstructed outputs. We compared the original and reconstructed EEG quantitatively and qualitatively.

Our results depicted in Fig. \ref{fig:fig2} described the proficiency of our masked autoencoder model in reconstructing EEG signals across various sleep stages. An observation revealed that the model adeptly managed to capture and replicated the characteristics inherent to each sleep stage \cite{webb1970sleep}. For instance, in the Wakefulness stage, the model effectively mimicked the original EEG signals, particularly in replicating the instances of relatively large amplitudes, exhibiting its sensitivity to the signal variations typical of this stage. In contrast, during the N1 state, where theta wave activation and reduced amplitudes were prevalent, the model's reconstructed signals reflected these characteristics, underscoring its adaptability and precision. This trend of accuracy and reliable replication continues across all sleep stages, including N2, N3, and REM, where the model proved its mettle by accurately reflecting essential features such as amplitude variations, low-frequency components enhancement, and the subtleties distinguishing REM from wakefulness. Such reconstruction accuracy across varied sleep stages underscores the model's robustness and reliability in EEG signal reconstruction, marking a significant stride in sleep study analyses.

The utilization of two primary signals, EEG and EOG, as inputs laid a foundational bedrock for a multifaceted analysis, while the calculation of MSE against a variety of reconstructed signals, barring the exclusion of self-reconstruction, adds a layer of depth and comprehensiveness to the evaluation process \cite{el1999validity}. A closer inspection of the tabulated MSE values revealed insightful patterns; for instance, when EEG served as the input, the model, with varying degrees of MSE, exhibited a discernable accuracy in reconstructing a multitude of signals such as EOG, EMG, and EEG across different sleep stages \cite{kweon2023development}. Similar trends of accuracy, reflected through the MSE values, are observed when EOG was employed as the input, reinforcing the model’s consistent performance and adaptability. This detailed tabulation of MSE values across a spectrum of scenarios provided a robust and nuanced evaluation platform, enabling a thorough assessment of the model's efficacy and reliability in the reconstruction of polysomnographic signals.

Although we proposed the masked autoencoder to reconstruct multi-signal from a single EEG channel and presented its performance using the Sleep-EDF-20 dataset, there were limitations to overcome for the expansion of PSG's applicability. First, the scope of our study was somewhat constrained by the utilization of a limited dataset, both in terms of the number of signals and participants included. This limitation underscored the necessity for subsequent research efforts to incorporate larger and more diverse datasets, which could potentially enhance the robustness and generalizability of the model \cite{ref8}. Second, the qualitative aspect of our comparison was not exhaustive. It did not encapsulate every nuanced feature intrinsic to the various sleep stages, leaving room for a more comprehensive and detailed qualitative analysis in future iterations of the study. This would allow for a richer and more insightful evaluation of the model's performance in capturing and replicating the subtleties of sleep stage characteristics. Finally, our study highlighted the need for a broader spectrum of metrics to facilitate a multi-dimensional comparison between the original and reconstructed signals. A more diverse array of evaluative metrics would enable a holistic assessment of the model's reconstruction accuracy, providing varied perspectives and insights into its performance and areas for improvement. Such an expanded evaluative approach would pave the way for a more nuanced understanding and optimization of the model’s capabilities in the reconstruction of polysomnographic signals.

\section{CONCLUSION}
In this study, we designed a masked autoencoder with the capability to reconstruct multi-signals using single-channel EEG data collected via PSG. Our model could reconstruct the multi-signals using only a single EEG channel by employing the MSE as a metric to evaluate the similarity between the original and the masked autoencoder's reconstructed outputs. Our results presented the potential to more simply diagnose sleep disorders and expand the applicability of PSG in a variety of settings by reconstructing a multi-signal from a single EEG channel.

\bibliographystyle{IEEEtran}
\bibliography{REFERENCE}

\end{document}